%% file: 0_Abstrct.tex
\documentclass[10pt,twocolumn,letterpaper]{article}
\usepackage[accsupp]{axessibility}  
\usepackage{wacv}
\usepackage{times}
\usepackage{epsfig}
\usepackage{graphicx}
\usepackage{amsmath}
\usepackage{amssymb}
\usepackage{booktabs}
\usepackage{ wasysym }
\usepackage{multirow}
\usepackage{multicol}
\usepackage{caption}
\usepackage{float}
\usepackage{xcolor}

\usepackage{caption}
\usepackage{algorithmicx,algorithm}
\definecolor{commentcolor}{RGB}{110,154,155}   
\definecolor{functioncolor}{RGB}{200,2,127}   

\def\wacvPaperID{1171} 

\wacvapplicationstrack 

\wacvfinalcopy 
\ifwacvfinal
\usepackage[breaklinks=true,bookmarks=false]{hyperref}
\else
\usepackage[pagebackref=true,breaklinks=true,colorlinks,bookmarks=false]{hyperref}
\fi

\begin{document}

\title{Few-shot Medical Image Segmentation with Cycle-resemblance Attention}

\author{Hao Ding$^{1}$, Changchang Sun$^{1}$, Hao Tang$^{2}$, Dawen Cai$^{3}$, Yan Yan$^{1}$ \\
$^{1}$Department of Computer Science, Illinois Institute of Technology, Chicago, IL, USA\\
$^{2}$Computer Vision Lab, ETH, Zurich, Switzerland \\
$^{3}$Department of Cell and Developmental Biology, University of Michigan, Ann Arbor, MI, USA\\
{\tt\small \{hding9,csun39\}@hawk.iit.edu, hao.tang@vision.ee.ethz.ch, dwcai@umich.edu, yyan34@iit.edu}
}

\maketitle
\thispagestyle{empty}

\begin{abstract}
    Recently, due to the increasing requirements of medical imaging applications and the professional requirements of annotating medical images, few-shot learning has gained increasing attention in the medical image semantic segmentation field. 
    To perform segmentation with limited number of labeled medical images, 
    most existing studies use Prototypical Networks (PN) and have obtained compelling success.
    However, these approaches overlook the query image features extracted from the proposed representation network, failing to preserving the spatial connection between query and support images.
    In this paper, we propose a novel self-supervised few-shot medical image segmentation network and introduce a novel Cycle-Resemblance Attention (CRA) module to fully leverage the pixel-wise relation between query and support medical images. 
    Notably, we first line up multiple attention blocks to refine 
    more abundant relation information. Then, we present CRAPNet by integrating the CRA module with a classic prototype network, where pixel-wise relations between query and support features are well recaptured for segmentation. Extensive experiments on two different medical image datasets, e.g., abdomen MRI and abdomen CT, demonstrate the superiority of our model over existing state-of-the-art methods.
 
\end{abstract}

\input{1_Introduction.tex}
\input{2_Related_work.tex}
\input{3_Proposed_Method.tex}
\input{4_Experiments.tex}
\input{5_Conclusion.tex}

{\small
\bibliographystyle{ieee_fullname}
\bibliography{egbib}
}


\end{document}

%% file: 1_Introduction.tex
\section{Introduction}

\begin{figure}[t]
\centering
\includegraphics[width=\linewidth]{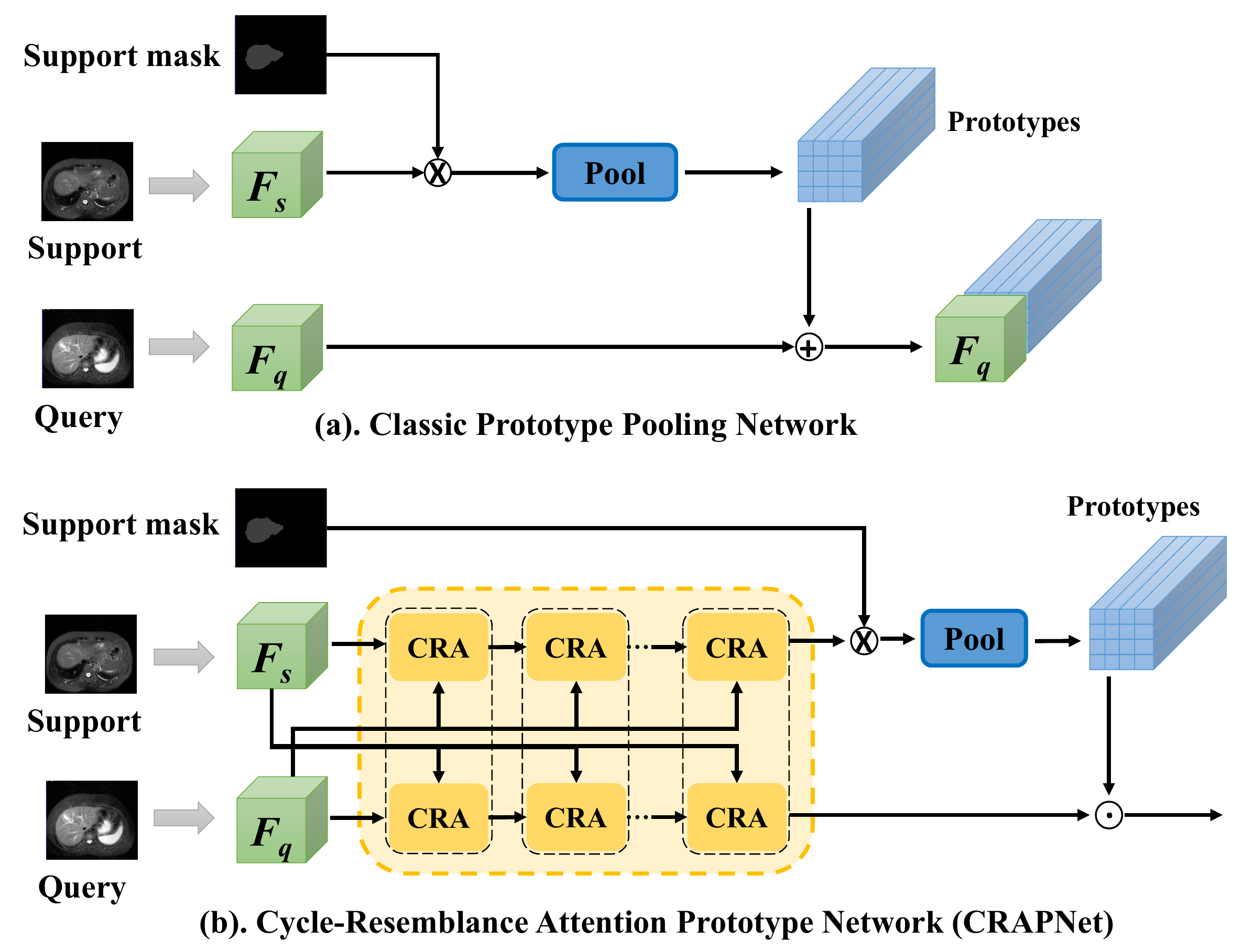}
   \caption{(a) The classical prototype pooling network. Prototypes are generated by pooling windows from extracting support features. 
   (b) Our proposed cycle-resemblance attention (CRA) module is plugged in front of the pooling step, where support and query features are integrated with each other via attention in a pixel level to enhance the spatial relationship between them. Furthermore, prototypes are introduced to guide the prediction of the query mask.
   }
\label{networkcompare}
\vspace{-0.4cm}
\end{figure}

Semantic segmentation is a fundamental task in computer vision and has achieved compelling success recently thanks to the flourishing of annotated data. Accordingly, it initiates the emerging real-world application of medical image segmentation, which can facilitate doctors for quicker disease diagnosis, better treatment planning, and treatment delivery.
To handle large-scale medical image efficiently, the accurate and professional label annotations are extremely important, unlike general images. However, it is pretty time-consuming and knowledge-required to annotate such a large amount of data \cite{milletari2016v, deselaers2008automatic, ko2012automatic, lutnick2019integrated, budd2021survey}.
Thus, in the medical imaging field, 
few-shot learning \cite{snell2017prototypical, tang2021recurrent, li2021adaptive, zhang2021self} has gained increasing attention from researchers due to its remarkable advantages of not requiring so much labeled data. Specifically, the discriminative representations can be extracted from one or a few pixel-wise annotated examples (support data) to realize the pixel-wise label prediction of unannotated samples (query data). 
Besides, compared to general images that are stored in the 2D format, medical images typically are highly structured 3D images of human organs and torso area, which have multiple forms, \textit{e.g.}, MRI (Magnetic Resonance Imaging), US (Ultrasound), CT (Computated Tomography) and X-ray \cite{bankman2008handbook, razzak2018deep, mcauliffe2001medical, semmlow2008biosignal, eklund2013medical, tournier2019mrtrix3, lehmann1999survey}.
The region of interest in medical images is usually tiny and homogeneous, while the irrelevant background is quite extensive and inhomogeneous \cite{wang2020self, tang2021recurrent}. 
Plentiful small cells, tissues, and organs tend to be squeezed together in  medical images, making it difficult to draw boundaries between foreground and background. 
    According to the mode of generating a prediction binary mask,
     existing few-shot image segmentation techniques can be broadly categorized into affinity learning, and prototypical learning \cite{li2021adaptive}. 
     The latter 
     designs the prototypical networks~\cite{snell2017prototypical,liu2022dynamic,zhang2021self,tang2021recurrent,wang2019panet,li2021adaptive} and generates prototypes that are generalized and robust to the noise.
     As shown in Figure~\ref{networkcompare}(a), the support image features are refined by the support mask and fed into a pooling module to obtain prototypes. Lastly, the prototypes are incorporated with query features adopting the plain operation, \textit{e.g.}, concatenation.
      Despite the promising performance of prototype-based methods, there are still several drawbacks. 
      (i) These methods inevitably lose the spatial information of support images, especially
      when the object appearance between the support and query images has large variation~\cite{li2021adaptive} due to excessive or insufficient amount of prototypes caused by imbalance object size between support and query images.
      (ii) The relationship between different classes in the images is critical in making segmentation decisions on query images, while current methods ignore it.
    (iii) In the training phase, current prototypical networks do not pay enough attention to the interaction between support features and query features. Such insufficient interaction would lead to the failure of generating a fully representative prototype. Nevertheless, due to the fact that 
    query images and support images share more similarities in both foreground and background, such interaction is vital in the task of image segmentation. Particularly, in the context of medical images, the arrangement of different objects usually follows resembling patterns between query and support images.

    To address issues
    as mentioned earlier,
    in this paper, we propose a novel few-shot medical image segmentation method with a cycle-resemblance attention mechanism, as shown in Figure~\ref{networkcompare}(b). Mainly, we introduce a new Cycle-Resemblance Attention Prototype Network (CRAPNet) to capture intrinsic object details fully and preserve spatial information between pixels inside query images and support images.
    As shown in Figure~\ref{pipeline},
     instead of giving an additive bias $B$ for the matched cycle-consistent pixel pairs via checking if they belong to the same class, we compare the similarity between those pixel pairs.
     In this way, a support-query-support connection is built, and we incorporate the relation between the pixel and its most similar ``neighbors" to obtain the prototypes.
    Moreover, inspired by looking deep down into the difference between support and query medical images, we argue that query and support images can be specially regarded as an interrupted video sequence or image flow, given the objects that are highly structured and organized. Therefore, we design the \textit{non-local} operations 
    that
    the cycle-resemblance module computes the weighted sum at a given pixel location for both support and query features with the non-local structure. In a sense, this non-local structure can be packed into a network block, which can be chained together and utilized as a \mbox{drop-in} module. Subsequently, the support and query branches are designed based on the aforementioned module, where the connection between them can be interactively characterized.

    The paper's contributions can be summarized as follows:
    \begin{itemize} 
    
        \item To the best of our knowledge, this is the first attempt to tackle the medical image segmentation task by designing a \mbox{Cycle-Resemblance} Attention Prototype Network (CRAPNet), which can preserve the spatial correlation between image features and smoothly incorporate it into the conventional prototype network. 
        \item A new non-local block with a built-in cycle-resemblance module is proposed, which can be chained together and utilized as a drop-in module. 
        \item Extensive experiments on two different medical imaging datasets, \textit{e.g.},
        abdomen MRI and abdomen CT, show the effectiveness of our proposed method.
    \end{itemize}
    

%% file: 2_Related_work.tex
\section{Related Work}
\subsection{Medical Image Segmentation}
With the development of computing hardware, deep learning approaches have stepped into the computer vision realm and started to show 
their powerful capability in image processing tasks \cite{hesamian2019deep}. In recent years,
deep neural networks have been booming significantly in medical image segmentation. 
In order to segment bones and tumors from the background, CNN-based network \cite{kayalibay2017cnn} is trained and tested on the human body and brain MRI. Later, Fully Convolutional Network (FCN) \cite{long2015fully} replaces the last fully connected layer in CNN with a fully convolutional layer which allows the network to have a dense pixel-wise prediction \cite{hesamian2019deep}.
For example, Roth \textit{et al.}~\cite{roth2018application} proposed a two-stage, coarse-to-fine approach using two chained 3D FCN networks, where the second network focused on more detailed segmentation of the organs and vessels. 
Besides, inspired by FCN \cite{long2015fully}, Ronneberger \textit{et al.}~\cite{ronneberger2015u} presented a well-known network for medical image segmentation task, named U-Net.
It is built upon FCN with a large number of feature channels during upsampling, allowing the network to propagate context information to higher resolution layers. 
Moreover,
Milletari \textit{et al.}~\cite{milletari2016v} re-designed the skip pathways of U-Net. The feature maps of the encoder are fed into a dense convolution block
instead of directly being input to the decoder. 
Such an operation is suitable 
when the feature maps from the decoder and encoder networks are semantically similar. 
Different from the above methods, 
V-Net \cite{milletari2016v} is another famous network in medical image segmentation, 
whose inputs are image volumes rather than the slices.
Particularly,
it demonstrates the fast and accurate results on 3D medical image volumes. 
To address the issue that current networks are limited to specific image analysis tasks, 
Isensee \textit{et al.}~\cite{isensee2021nnu} presented the nnU-Net framework, which combines three simple U-Net and automatically adapts its network architectures to the given image geometry. 
However, these approaches all require a large amount of annotated data in order to enable their full potentials, and lack the ability to make segmentation predictions on new classes. 


\subsection{Few-shot Segmentation}

Due to the lack of annotated data, Few-shot Semantic Segmentation (FSS) techniques have been wildly explored recently. For example, Shaban \textit{et al.}~\cite{shaban2017one} put forward a novel two-branched approach to a one-shot segmentation task, which is the first to address few-shot semantic segmentation. 
In particular, the first branch takes a labeled image as input and outputs a parameterized vector. The second branch takes the vector and another image as input and outputs the segmentation mask of the image for a new class~\cite{shaban2017one}.
Inspired by the fact that embeddings of each point cluster can represent the prototype of the corresponding category, Snell \textit{et al.} \cite{snell2017prototypical} established the \textit{Prototypical Networks}, which can handle both the few-shot and zero-shot settings.
 Specifically,
 few-shot prototypes are computed using the average of support examples of different classes,
 while zero-shot prototypes are calculated on a high-level description of the classes that come with the meta-data. 
 The framework is pretty intuitive and straightforward, but it achieves a significant performance. 
 From then on, 
 more attention has been paid to the prototype-based methods.
 For instance,
 SG-One \cite{zhang2020sg}
 works on leveraging the support image mask and adopting an average pooling module to preserve class-specific information while generating prototypes.
 To make segmentation predictions, cosine similarity between prototypes of classes and query image features is utilized.
 Besides,
to fully exploit the support knowledge and improve the few-shot learning performance,
PANet \cite{wang2019panet} exploits metric learning and introduces a novel prototype alignment regularization strategy.
In addition, based on the image content, ASGNet \cite{li2021adaptive} uses a Guided Prototype Allocation (GPA) strategy to adaptively decide the number of prototypes and their spatial extent.
Furthermore,
DPCN \cite{liu2022dynamic} introduces a dynamic convolution module to achieve adequate support-query interactions along with a support activation module (SAM).
Moreover, a Feature Filtering Module (FFM) is appended to mine the complementary information from query images. 
Despite of the compelling success achieved by these methods in
general cases, little attention has been paid to the bias problem.
Therefore,
Lang \textit{et al.} \cite{lang2022learning} 
 proposed to estimate the scene differences between the query-support image pairs through the Gram matrix, so as to mitigate the adverse effects caused by the sensitivity of meta learner. All these approaches have their limitation in terms of support-query connection before acquiring prototypes. Thus, in this work, we focus on leveraging spatial information with such connection to aid prototype generation and show that it benefits the task of few-shot semantic segmentation. 
 






\begin{figure*}
\centering
\includegraphics[width=0.95\textwidth]{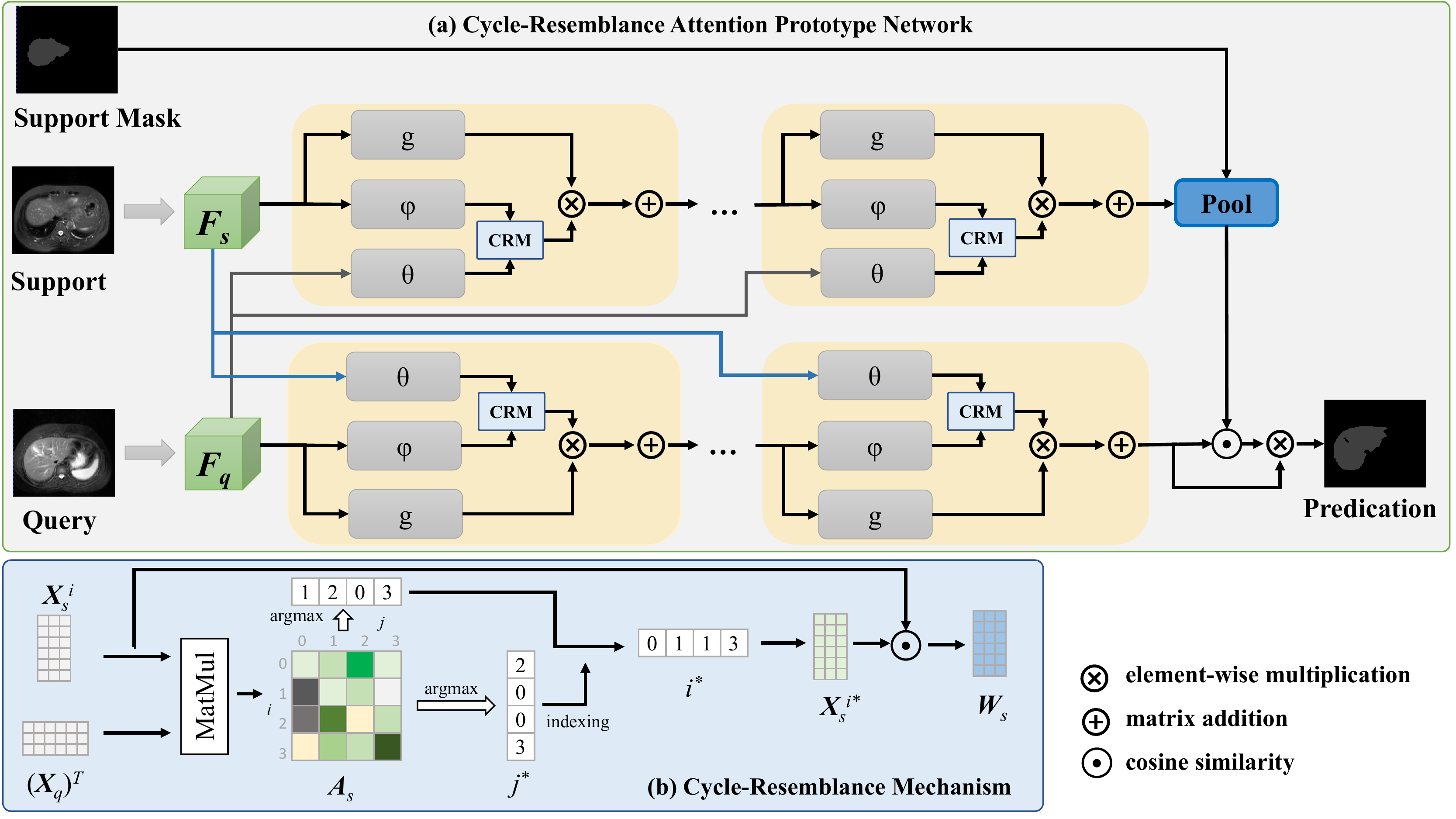}
   \caption{(a) The extracted features from backbone network are first input to 5 support-query attention block in each branch, 
   where attention blocks $g, \varphi, \theta$ are $1\times 1\times 1$ convolution operation. The module denoted as \textbf{CRM} between $\theta$ and $\varphi$ exploits the cycle-resemblance mechanism. (b) Cycle-Resemblance first calculates matrix multiplication between support and query feature maps after $\varphi$ and $\sigma$ convolution. Then, for pixel $i$ in the support feature map, the most similar pixel $j^*$ is found in the query feature by looking up the matrix. 
     For $j^*$, the most similar pixel  $i^*$ is found as well.  
   Last, cosine similarity between features $\textbf{x}_s^i$ and $\textbf{x}_s^{i^*}$ is calculated, and a
   softmax function is adopted to return the weight for pixel $i$.
   }
\label{pipeline}
\end{figure*}

%% file: 3_Proposed_Method.tex
\section{Method}

\subsection{Problem Definition}

In the context of few-shot segmentation,
the dataset is split into
two parts, training dataset $D_{train}$ and testing dataset $D_{test}$. 
Both datasets consist of image-binary mask pairs,
and
$D_{train}$ is annotated by $L_{train}$, and  $D_{test}$ is annotated by $L_{test}$.
Intuitively, 
there are no common classes between two class sets, \textit{i.e.,} $L_{train} \cap L_{test} = \varnothing$.
Following the problem setting of the initial few-shot semantic segmentation work \cite{shaban2017one}, suppose that we have 
support image set $S = \{(I^i_s, Y^i_s(l)), l\in L_{test}\}_{i=1}^k$, where $I_s^i$ is the $i$-$th$ image in the support image set, $Y_s^i(l)$ is the mask of $i$-$th$ support image of class $l$.
The objective of few-shot semantic segmentation is to learn a function $f(I_q, S)$, which predicts a binary mask of an unseen class $I_q$
when given the query image $I_q$ and the support set $S$.

While training the few-shot networks, the input data of the model are $\langle S, I_q \rangle$ pairs. Specifically, $S$ is a subset of $D_{train}$ ($S \subset D_{train}$). 
$(I_q, Y_q(l))
\notin S$, and $Y_q(l)$ is only used for training.
We denote such pair as an episode, and 
each episode is randomly sampled from $D_{train}$.
There are totally $k$ image-binary mask pairs in support set $S$ for the semantic class $l$, and $n$ classes in $Y_s$. Thus, it is called an $n$-way $k$-shot segmentation sub-problem for each episode.

\subsection{Network Overview}

We use the self-supervision framework~\cite{ouyang2020self} for our few-shot semantic segmentation task. 
Our method includes three components:
1) A feature encoder to extract feature representation from input medical image; 2) A generic support-query attention encoder to encode both support and query features supported by cycle-Resemblance mechanism; 3) A similarity-based classifier for segmentation conditioned on support prototype and decoded query feature.
\subsection{Feature Extraction}\label{featureextraction}

As we discussed above, the unit of input for the few-shot model is an episode $\langle S, I_q\rangle$, and we use a feature extraction encoder $f_{\theta}$ to extract both support features $f_{\theta}(I_s^i) \in \mathbb{R}^{D\times H\times W}$ and query features $f_{\theta}(I_q) \in \mathbb{R}^{D\times H\times W}$
parameterized by $\theta$. $H$ and $W$ are the height and width of feature maps, and $D$ represents the channel depth. 
Specifically, the feature extraction encoder takes an input image in the dimension of $3\times 256 \times 256$ and outputs a $256\times 32\times 32$ feature map.
Then,
we use fully-conventional ResNet-101 backbone as our feature encoder, which is specifically pre-trained on the part of MS-COCO for better segmentation performance \cite{shin2016deep, wang2019panet, ouyang2020self}.
In practice, 
we adopt the \texttt{deeplabv3\_resnet101} model from ``torchvision'' python library.

\subsection{Support-query Attention Module}\label{att}
Different from prior works, which directly generate prototypes from support feature maps with the help of support masks and compare prototypes with query features to the classification, we design a support-attention module to concern 
the connections and relationships between support and query features.
Specifically, we use non-local mean operation \cite{buades2005non, wang2018non}, and construct a similar network structure. 
For the obtained support and query feature maps,
we define the support-query attention encoder as follows,
\vspace{-0.05in}
\begin{equation}
    \textbf{y}_s = \dfrac{1}{\mathcal{C}(\textbf{x})} \sum_{\forall i} f(\textbf{x}_q, \textbf{x}_s^i)g(\textbf{x}_s^i),
    \label{eq1}
\end{equation}
\vspace{-0.1in}
\begin{equation}
    \textbf{y}_q = \dfrac{1}{\mathcal{C}(\textbf{x})} \sum_{\forall i} f(\textbf{x}_s^i, \textbf{x}_q)g(\textbf{x}_q),
    \label{eq2}
\end{equation}
where $\textbf{x}_s^i$ is the support image feature extracted from the $i$-$th$ image of support set $S$
by above mentioned 
feature extraction encoder $f_{\theta}$.
Similarly, $\textbf{x}_q$ is the query feature.
Both can be obtained as follows,
\vspace{-0.05in}
\begin{equation}
    \textbf{x}_s^i = f_{\theta}(I_s^i),
\end{equation}
\vspace{-0.2in}
\begin{equation}
    \textbf{x}_q = f_{\theta}(I_q).
\end{equation}
Concretely, the extracted support features $\textbf{x}_s^i$ and query feature $\textbf{x}_q$ are first flattened in terms of pixels, \textit{e.g.}, $\textbf{x}_s^i \in  \mathbb{R}^{D\times HW}$ and $\textbf{x}_q \in \mathbb{R}^{D\times HW}$, and then each is applied with the $1\times 1\times 1$ convolution. 
Moreover,
$f$ is the pair-wise function that computes a weighted map representing pixel-wise relationships between support and query feature maps. This function is provided by the cycle-resemblance mechanism, which
will be explained in Sec.~\ref{cyc}.
$g$ is a function that computes the representation of input support or query feature maps. $\mathcal{C}(\textbf{x})$ is a normalization factor, where $\textbf{x}$ is the concatenation of input support features and query feature. 
The advantage of such 
non-local operation is that all image features are considered together to obtain learned weight maps and achieve a representation of both support and query branches. Specifically,
function $g$ is designed as linear embeddings as follows,

\begin{equation}
   g(\textbf{x}_s^i) = {\mathbf{W}_s} \textbf{x}_s^i,
\end{equation}
\vspace{-0.2in}
\begin{equation}
    g(\textbf{x}_q) = \textbf{W}_q \textbf{x}_q,
\end{equation}
where $\textbf{W}_s$, $\textbf{W}_q$ are weight maps learned from pairwise function $f$ provided by the cycle-resemblance mechanism, representing the inner spatial information of either query or support features.
Finally, we compress the non-local operation into a block, which can be defined as,

\begin{equation} \label{eq5}
    \textbf{z}_s = \textbf{W}_s\textbf{y}_s + \textbf{x}_s,
\end{equation}
\vspace{-0.2in}
\begin{equation}
    \textbf{z}_q = \textbf{W}_q\textbf{y}_q + \textbf{x}_q,
\end{equation}

\noindent where $\textbf{y}_s$ and $\textbf{y}_q$ are obtained by Eq.~\eqref{eq1} and Eq. \eqref{eq2}, respectively.
$\textbf{x}_s$ is the feature set of all the support images.
The ``+" 
denotes a residual connection \cite{he2016deep, wang2018non}, which allows the block to be applied after the feature 
extraction without breaking any initial behavior. Thus, this block serves a plug-and-play functionality and can be easily deployed.

\subsection{Cycle-resemblance Mechanism} \label{cyc}
As mentioned in Sec.~\ref{att}, the cycle-resemblance module is responsible for building an inner pixel-wise relationship
when the interaction between support and query features maps. 
Specifically, we turn to the cycle-consistent strategy~\cite{zhang2021few} and obtain the weighted maps by applying them to the support and query feature maps.
Formally, $\mathbf{A}_s$ and $\mathbf{A}_q \in \mathbb{R}^{HW\times HW}$
are calculated via matrix multiplication to represent the similarity between flattened support feature $\textbf{x}_s^i(l)$ and query feature $\textbf{x}_q$ . For the support branch,

\begin{equation}
    \mathbf{A}_s= \textbf{x}_s^i(\textbf{x}_q)^T,
\end{equation}
and for the query branch,

\begin{equation}
    \mathbf{A}_q = \textbf{x}_q (\textbf{x}_s^i)^T.
\end{equation}
For simplicity, 
we take the computation in the support branch 
as an example, and that of the query branch can be
effortlessly achieved in the same manner. We use $\mathbf{A}$ to represent $\mathbf{A}_s$ for the following illustration.
For the single pixel at the position $j$ ($j \in {0, 1, ..., HW}$) of the 
support feature map, its most similar point $i^*$ in the query feature can be acquired by finding the index whose value is maximized along the column direction, which can be denoted as,

\begin{equation}
    i^* = \underset{i}{\text{argmax }} \mathbf{A}_{(i,j)}.
\end{equation}
Obtaining the most similar pixels 
of each pixel in support features
from query features,
we can perform the mapping from the support feature to the query features.
Specifically,
for the point $i^*$, we have 
\begin{equation}
    j^* = \underset{j}{\text{argmax }} \mathbf{A}_{(i^*,j)}.
\end{equation}
Different from the work \cite{zhang2021few}, which establishes a cycle-consistency
mechanism and compute the binary value (0 or 1) of the support mask at pixel location $j$ and $j^*$ by checking if
$Y_s^i(l)_{j}$ equals to $Y_s^i(l)_{j^*}$,
in our approach, 
we design a resemblance weight map without involving any labels 
and 
introduce a novel cycle-similarity operation to 
maximize the 
resemblance between support and query features which are the most relevant, and we obtain the pair-wise function $f$
by defining the weighted map $\textbf{W}_s$ as follows,
\begin{equation}
    \textbf{W}_s = \text{softmax}\big(\textbf{x}_s^i(j) \text{ }\astrosun \textbf{ x}_s^i(j^*)\big),
\end{equation}
where $\astrosun$ denotes the cosine similarity operation, and $\textbf{ x}_s^i(j)$ and $\textbf{ x}_s^i(j^*)$ are the features at pixel location $j$ and $j^*$ from support feature map, respectively. 



\subsection{Prototype and Similarity-based Segmentation}

To obtain the global prototype of all the classes, 
we use CARAFE++ \cite{wang2019carafe, wang2021carafe++}, 
a content-aware feature reassemble technique, to extract prototypes $p$ from $\mathbf{z}_s$, which is obtained from Eq.~\eqref{eq5}. 
Therefore, the prototype at location $(m, n)$ is 
denoted as,

\begin{equation}
    p_{(m,n)} = \sum_{i=-r}^r \sum_{j=-r}^r {\textbf{W}_{(h', w')}}(i, j) \cdot \textbf{z}_s(h+i,w+j), 
\end{equation}
where $\textbf{W}_{(h',w')}(i, j)$ is a predicted location-wise kernel for location $(h', w')$ based on the neighbors of $\textbf{z}_s(h+i, w+j)$. $i$, $j$ $\in{[-r,r]}$ 
, standing
for the searching range when finding the neighbors of $(h, w)$.
$h'=\lfloor h/ \sigma \rfloor$, and $w'=\lfloor w/ \sigma \rfloor$, giving $\sigma$ as the downsampling factor. and $r$ equals to half of the reassembly kernel size. 
Additionally, we compute class-level prototype $p^i(l^{\hat j})$ with the engagement of support mask 
$Y_s^i$ \cite{ouyang2020self, wang2019panet, zhang2020sg} as follows,
\begin{equation}
    p^i(l^{\hat j}) = \dfrac{\sum_h \sum_w Y_s^i(l^{\hat j}) \textbf{x}_s^i(h,w)}{\sum_h \sum_w Y_s^i(l^{\hat j})(h, w)},
\end{equation}
where $\hat j$ refers to the classes except for the background class. 
Then, the prototype of each class can be formed together via concatenation operation, termed as $P = \{p(l^j)\}$.


For the similarity-based classifier, the target is to make a dense prediction of query conditioned on support prototypes \cite{ouyang2020self}. In our work, the similarity of 
$j$-$th$
class at pixel location $(h, w)$ is defined as,

\begin{equation}
    S_{l^j}(h,w)=\alpha p(l^j)\textbf{ } \astrosun \textbf{ }\textbf{x}_q(h,w),
\end{equation}
where $\alpha$ is a multiplier,
serving as a constant to 
assist 
the gradient backpropagation \cite{ouyang2020self, oreshkin2018tadam}. Similar to \cite{ouyang2020self, wang2019panet}, 
the value of $\alpha$ is set to be 20 in our work.
Finally,
to predict the pixel-wise class $\hat Y_q(h,w)$, we apply the softmax function to  cosine-similarity maps as follows,
\begin{equation}
    \hat Y_q(h,w) = \text{softmax}\bigg( S_{l^j}(h,w)\cdot\text{softmax}\big( S_{l^j}(h,w)\big)\bigg).
\end{equation}

\subsection{Loss Function}

Superpixel pseudo-label, which contains rich clustering information, is a good replacement when the annotation is absent.
Intuitively, a semantic mask consists of several superpixels \cite{ren2003learning, stutz2018superpixels, ouyang2020self}.
A series of superpixels $Y_i$ are generated through function $\mathcal{F}$ for each image $I^i$ in the dataset, 
denoted as $Y^i(l^p) = \mathcal{F}(I^i)$, where $i$ means the $i$-$th$ image, and $l^p$ represents the pseudo-label class. In addition, the background of corresponding pseudo-labels is defined by $Y^i(l^0) = 1-Y^i(l^p)$. The superpixel and its corresponding image 
are
randomly picked from the support set. The query image is formed by $\mathcal{T}_g(\mathcal{T}_{\gamma}(I^i))$, where $\mathcal{T}_g$ denotes the affine and elastic transform, and $\mathcal{T}_{\gamma}$ is gamma transformation. However, for pseudo-labels of query images, only geometry transform is applied, \textit{e.g.}, $\mathcal{T}_g(Y^i(l^p))$.

In this paper, we adopt 1-way 1-shot approach for our segmentation task, and our loss function consists of two parts, segmentation loss \cite{ouyang2020self} and prototypical alignment regularization loss \cite{wang2019panet}.
In each iteration $t$, an episode of support query image pair $\langle S_t, Q_t\rangle$ is taken as the input, and then the segmentation loss is computed via cross-entropy loss as follows, 

\begin{equation}
\begin{split}
        & \mathcal{L}_{seg}^t (\theta; S_t, Q_t) =  \\ & -\dfrac{1}{HW}  \displaystyle\sum^H_h \sum_w^W \sum_{j\in {0, p}}\mathcal{T}_g\big(Y_t(l^j)\big)(h,w) \cdot \log\big(\hat{Y}_t(l^j)(h,w)\big),
\end{split}
\end{equation}
where $\hat{Y}_t(l^j)$ is the predicted results of pseudo-label $\mathcal{T}_g(Y_t(l^j))$. 
We use the same weight factor for cross-entropy loss, where $0.05$ for background class $l^0$ and $1.0$ for the other class $l^p$.
Then, we employ the prediction result along with corresponding image as the support $S' =  \{\mathcal{T}_g(\mathcal{T}_{\sigma}(I_t), \hat{Y}_t(l^j))\}$.
Thereafter,
the regularization loss can be formed as follows,
\begin{equation}
    \begin{split}
        & \mathcal{L}_{reg}^t (\theta; S'_t, S_t) = \\ & -\dfrac{1}{HW} \displaystyle\sum^H_h \sum_w^W \sum_{j\in {0, p}}Y_t(l^j)(h,w) \cdot \log\big(\bar{Y}_t(l^j)(h,w)\big).
\end{split}
\end{equation}
Above all,
we reach the final
objective formulation $\mathcal{L}^t(\theta; S_t, Q_t)$ as follows,
\begin{equation}
    \mathcal{L}^t(\theta; S_t, Q_t) =  \mathcal{L}_{seg}^t + \lambda \mathcal{L}_{reg}^t,
\end{equation}
where $\lambda$ is the 
nonnegative tradeoff parameter.

%% file: 4_Experiments.tex
\section{Experiments}

\begin{table*}[t]
\centering
\begin{tabular}{lcccccccccc}
\toprule
                        & \multicolumn{5}{c}{Abdominal-CT}                                                                           & \multicolumn{5}{c}{Abdominal-MRI}                                                                          \\
\multirow{2}{*}{Method} & \multicolumn{2}{c}{Kidneys}     & \multirow{2}{*}{Spleen} & \multirow{2}{*}{Liver} & \multirow{2}{*}{Mean} & \multicolumn{2}{c}{Kidneys}     & \multirow{2}{*}{Spleen} & \multirow{2}{*}{Liver} & \multirow{2}{*}{Mean} \\
                        & LK             & RK             &                         &                        &                       & LK             & RK             &                         &                        &                       \\ \midrule
ALPNet~\cite{ouyang2020self}                  & 29.12          & 31.32          & 41.00                   & 65.07                  & 41.63                 & 44.73          & 48.42          & 49.61                   & 62.35                  & 51.28                 \\
SSL-PANet~\cite{ouyang2020self}               & 56.52          & 50.42          & 55.72                   & 60.86                  & 57.88                 & 58.83          & 60.81          & 61.32                   & 71.73                  & 63.17                 \\
SSL-ALPNet~\cite{ouyang2020self}              & 72.36          & 71.81          & \textbf{70.96}          & \textbf{78.29}         & 73.35                 & 81.92          & 85.18          & 72.18                   & 76.10                  & 78.84                 \\
SSL-RPNet~\cite{tang2021recurrent}            &65.14           &66.73           &64.01                    &72.99                   & 67.22                 & 71.46          & 81.96          & 73.55                   & 75.99                  & 75.74 \\
CRAPNet (Ours)        & \textbf{74.69} & \textbf{74.18} & 70.37                   & 75.41                  & \textbf{73.66}        & \textbf{81.95} & \textbf{86.42} & \textbf{74.32}          & \textbf{76.46}         & \textbf{79.79}        \\ \bottomrule
\end{tabular}
\caption{Experimental results (in Dice Score) on abdominal images in \textit{\textbf{setting 1}}.}
\label{tab:setting1}
\vspace{-0.1cm}
\end{table*}

\begin{table*}[t]
\centering
\begin{tabular}{lcccccccccc}
\toprule
                        & \multicolumn{5}{c}{Abdominal-CT}                                                                       & \multicolumn{5}{c}{Abdominal-MRI}                                                                      \\
\multirow{2}{*}{Method} & \multicolumn{2}{c}{Kidneys} & \multirow{2}{*}{Spleen} & \multirow{2}{*}{Liver} & \multirow{2}{*}{Mean} & \multicolumn{2}{c}{Kidneys} & \multirow{2}{*}{Spleen} & \multirow{2}{*}{Liver} & \multirow{2}{*}{Mean} \\
                        & LK           & RK           &                         &                        &                       & LK           & RK           &                         &                        &                       \\ \midrule
ALPNet-init~\cite{ouyang2020self}             & 13.90        & 11.61        & 16.39                   & 41.71                  & 20.90                 & 19.28        & 14.93        & 23.76                   & 37.73                  & 23.93                 \\
ALPNet~\cite{ouyang2020self}                  & 34.96        & 30.40        & 27.73                   & 47.37                  & 35.11                 & 53.21        & 58.99        & 52.18                   & 37.32                  & 50.43                 \\
SSL-PANet \cite{ouyang2020self}               & 37.58        & 34.69        & 43.73                   & 61.71                  & 44.42                 & 47.71        & 47.95        & 58.73                   & 64.99                  & 54.85                 \\
SSL-ALPNet \cite{ouyang2020self}              & 63.34        & 54.82        & 60.25                   & \textbf{73.65}         & 63.02                 & 73.63        & 78.39        & 67.02                   & 73.05                  & 73.02                 \\
CRAPNet (Ours)        & \textbf{70.91}  & \textbf{67.33} & \textbf{70.17}     & 70.45                  & \textbf{69.72}        & \textbf{74.66} & \textbf{82.77} & \textbf{70.82} & \textbf{73.82} & \textbf{75.52}                       \\ \bottomrule
\end{tabular}
\caption{Experimental results (in Dice Score) on abdominal images in \textit{\textbf{setting 2}}.}
\label{tab:setting2}
\end{table*}

\subsection{Datasets}
To verify the generality and robustness of our approach, 
we perform semantic segmentation experiments on two datasets, abdominal MRI and abdominal CT image scans.
Abdominal CT is a dataset from MICCAI 2015 Multi-Atlas Abdomen Labeling challenge \cite{miccai2015}, which contains 30 3D abdominal CT scans of total 13 different labels. 
In our work, we only select four conjunct labels: left kidney, right kidney, liver, and spleen.
The abdominal MRI dataset is from Combined Healthy Abdominal Organ Segmentation (CHAOS) challenge \cite{kavur2021chaos} held in IEEE International Symposium on Biomedical Imaging (ISBI) 2019. It consists of 20 3D MRI scans with total four different labels. In order to apply 5-fold cross-validation as our evaluation approach, each dataset is partitioned into 5 parts evenly.

\subsection{Evaluation Metrics} \label{eva}
We adopt  dice score to gauge our segmentation model performance, ranging from 0 to 100, where $0$ stands for 
the pixel-wise overlap between segmentation prediction and ground truth is zero, while 100 means a 100\% perfectly match.
Besides,
to validate the generalization ability of our model on unseen labels, we employ the established few-shot segmentation experiment setting in \cite{roy2020squeeze} as ``setting 1", where the testing class may appear on the background of training images and
we train and test on all four labels without any partitioning.
Meanwhile,
we also follow the experiment setting in baseline method \cite{ouyang2020self} as ``setting 2", where the testing class do not appear in any training images.
In detail, due to the fact that left and right kidneys often appear simultaneously,
we group the left and right kidneys together in one label group, 
and put the spleen and liver together in another label group.



\subsection{Implementation Details}
We implement our network with Pytorch, and
we use fully conventional ResNet-101 \cite{DBLP:journals/corr/HeZRS15} as our feature extraction encoder $f_\theta$, which has been pre-trained on the part of MS-COCO as we mentioned in Sec. \ref{featureextraction}.
We scale the image into $3\times 256 \times 256$, and obtain the feature map in the shape of $256 \times 32 \times 32$ after the encoder. The downsampling factor $\sigma $ is set as $4$ for CARAFE++. Thus, the output prototype after CARAFE++ is $256\times 8 \times 8$. 
Moreover,
we use SGD as our optimization function with an initial learning rate of 0.001. We apply a multi-step learning rate scheduler to dynamically change our learning rate every 1000 iterations. 
Meanwhile, we set the batch size to 1, and the total iterations to 100,000. The model is trained on NVIDIA RTX A5000 (24G) GPU for about 4 hours per fold, and the memory consumption is around 7GB.

\begin{figure*}
\centering
\includegraphics[width=0.95\textwidth]{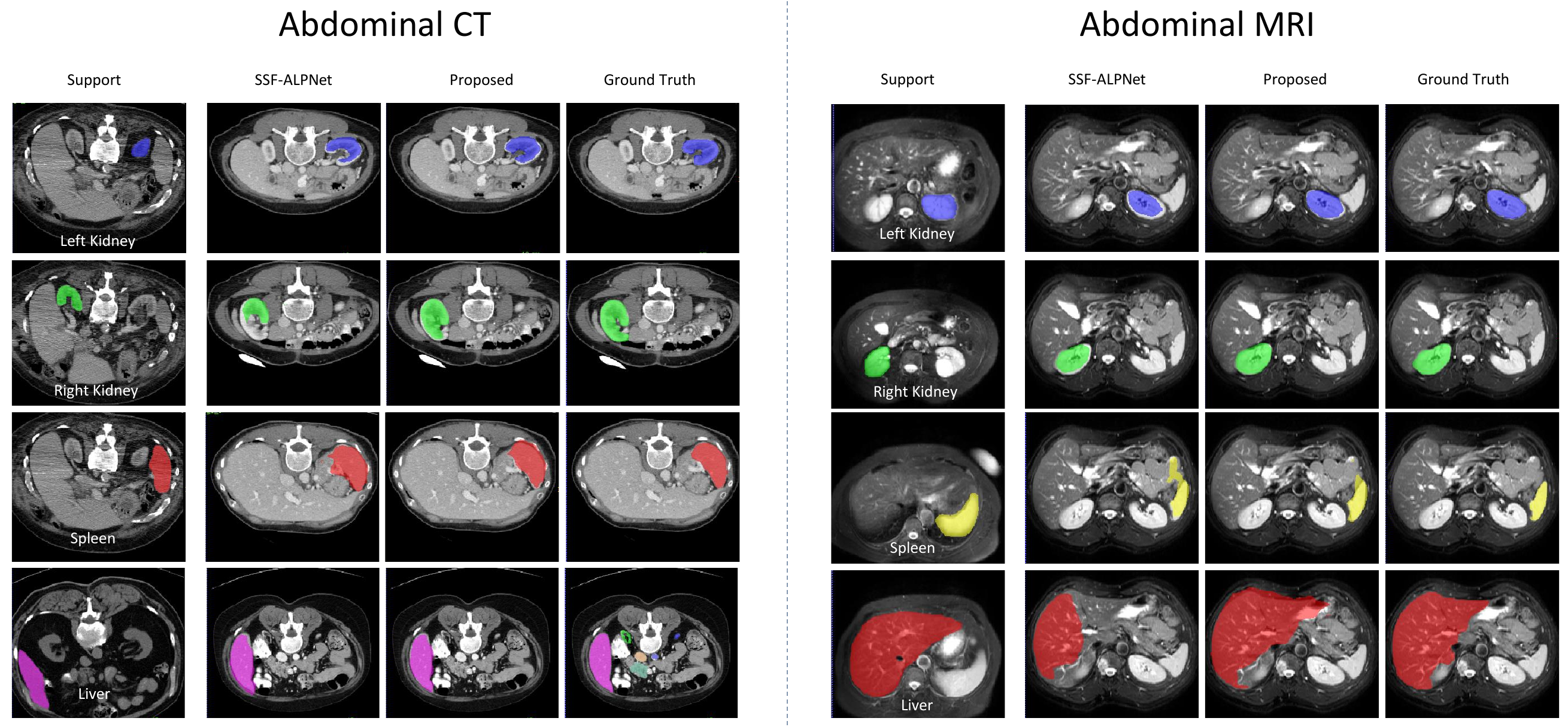}
\caption{The qualitative results of experiments in \textbf{\textit{setting 1}} on both datasets.}
\label{visualres}
\end{figure*}

\subsection{Quantitative and Qualitative Results}
To comprehensively evaluate the proposed CRAPNet,
we compare the proposed CRAPNet with several state-of-the-art medical image semantic segmentation baselines, including 
ALPNet, SSL-PANet, SSL-ALPNet, and SSL-RPNet. 
Compared with ALPNet, SSL-ALPNet introduces a superpixel-based self-supervision module to solve the lack of labels,
and SSL-PANet removes the adaptive pooling component.
RPNet is a supervised method that uses a recurrent mask refine module to iteratively refine the segmentation mask.
In order to compare the method with our method in the same self-supervision setting, we adapt it into our self-supervision learning framework, which is denoted as SSL-RPNet.
We first report the results under setting 1 in Table \ref{tab:setting1}, and then display the results under setting 2 in Table \ref{tab:setting2}.
From Tables~\ref{tab:setting1} and~\ref{tab:setting2}, 
we can draw the following observations:
1) Overall, Abdominal-CT dataset, the performance of CRAPNet is significantly
better than all baselines, except for the SSL-ALPNet with the Spleen and Liver categories.
2) Our CRAPNet consistently
outperforms all the other baselines with different organs on Abdominal-MR dataset, and has a better mean classification accuracy on the Abdominal-CT dataset.
In particular, on Abdominal-MR dataset,
compared with the best baseline, CRAPNet
achieves a significant average improvement of 0.95\%,
and 2.5\% in setting 1 and 2, respectively.
This can be attributed to the
fact that the image quality of the Abdominal-MR dataset is better than that of the Abdominal-CT dataset.



To gain more deep insights,
we also show the visual segmentation results with setting~1 in Figure \ref{visualres}.
As can be seen, our model makes a more precise segmentation especially on spleen and liver organ of the MRI dataset. 
Moreover, 
take the left kidney for example, 
our model obtains better prediction on the boundary of the left kidney.

\subsection{Ablation Study}
To verify the influence of the number of attention blocks and the effectiveness of the two branch attention blocks, we further conduct experiments on the abdominal-MRI dataset.


\begin{table}[]
\centering
\begin{tabular}{cccccc}
\toprule
\# of Blocks & LK    & RK   & Spleen         & Liver          & Mean           \\ \midrule
1                & 80.39          & 82.42          & \textbf{74.52}  & 71.93          & 77.30          \\
5                & 81.95          & \textbf{86.42} & 74.32 & \textbf{76.46} & \textbf{79.79} \\
7                & 82.08          & 83.93          & 73.35          & 73.16          & 78.13          \\
9                & 80.27          & 83.93          & 73.61          & 74.28          & 78.02          \\
12               & \textbf{82.41} & 85.84          & 71.88          & 73.02          & 78.29          \\
15               & 80.71          & 86.00          & 73.88          & 72.67          & 78.31          \\ \bottomrule
\end{tabular}
\caption{Experiments results (in Dice Score) on the number of Cyc-Resemblance blocks on abdominal MRI in \textit{\textbf{setting~1}}.}
\label{tab:numblocks}
\vspace{-0.3cm}
\end{table}

\noindent\textbf{Number of Attention Blocks.} 
The sensitivity analysis of the number of attention blocks is shown in Table~\ref{tab:numblocks}, where we vary the number from $1$ to $15$. As can be seen, 
the optimal performance can be achieved when the number is set as $5$, indicating that
stacking multiple attention blocks is beneficial. The excessive amount of blocks also causes a negative effect on the model, making the network pay too much attention to local details and thus ignore the global information.

\begin{figure}[htbp]
\centering
\includegraphics[width=0.95\linewidth]{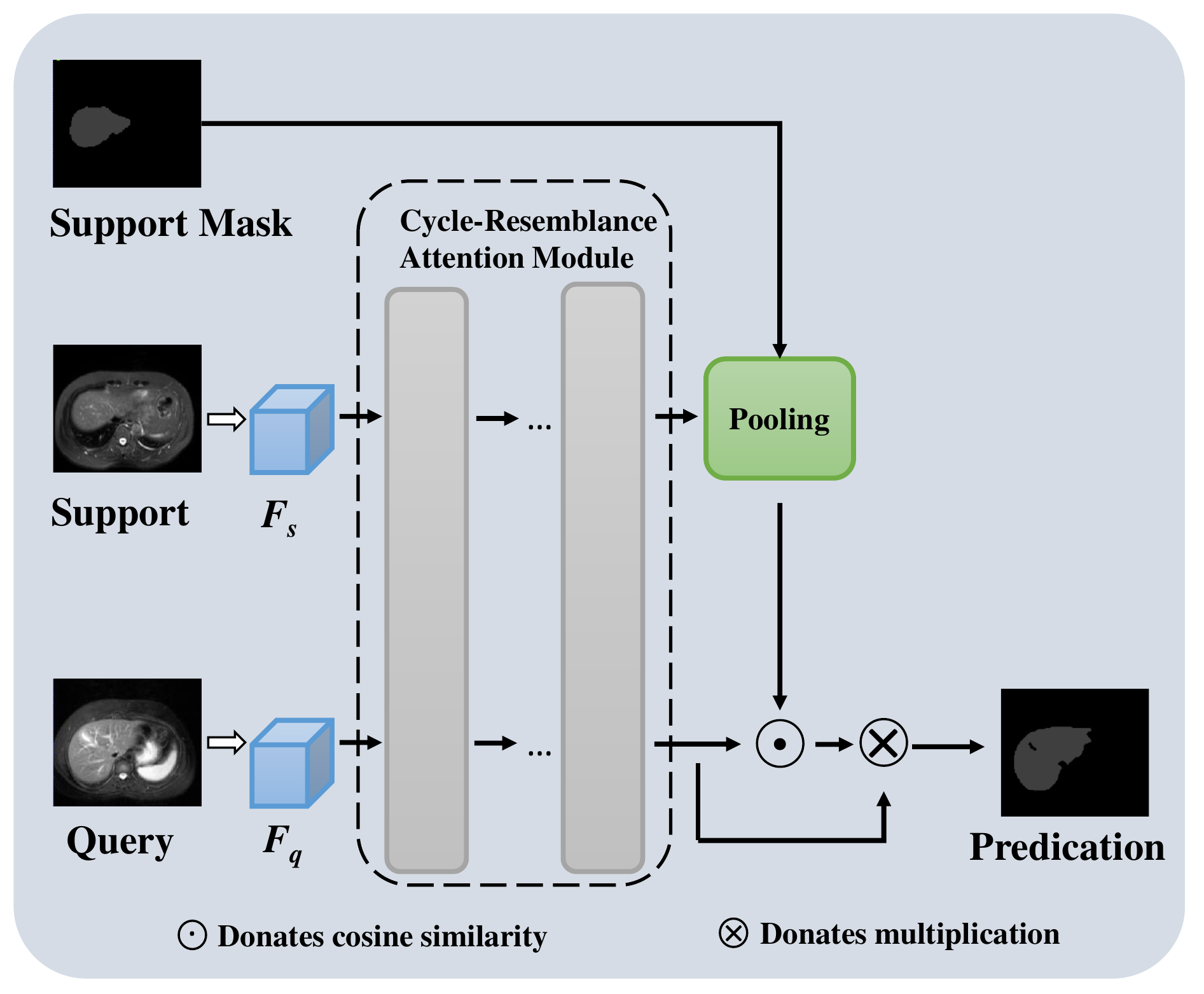}
   \caption{Single branch attention block.}
\label{another}
\vspace{-0.4cm}
\end{figure}

\noindent\textbf{Single Branch Attention Block.} 
Additionally, 
to better explain the benefit of introducing two-branch attention blocks, we conduct the comparative experiment with one derivative of our model only containing a single branch. 
Specifically, as shown in Figure~\ref{another}, 
it can be regarded as fusing the two branches into a single branch.
We force the support and query features to update simultaneously when they are input to the next block.
Table~\ref{tab:block2} shows the performance on the abdominal-MRI dataset.
As can be seen,
CRAPNet consistently has better performance no matter which organ category, and this 
 well validates the necessity of
taking two branches into account in the context of medical image segmentation.
In particular,
the overall dice score drops over $3$\%, and the potential reason is that 
some critical information is lost when updating the support and query features.
Nevertheless, when two branches are considered, there is only one kind of feature will be updated, with another one unchanged.



 

\begin{table}[]
\centering
\begin{tabular}{cccccc}
\toprule
Approach & LK    & RK   & Spleen         & Liver          & Mean           \\ \midrule
2-branch                & \textbf{81.95}          & \textbf{86.42} & \textbf{74.32} & \textbf{76.46} & \textbf{79.79} \\
1-branch               & 78.08          & 81.44         & 71.65          & 73.33          & 76.13          \\ \bottomrule
\end{tabular}
\caption{Experiments results (in Dice Score) on single branch implementation of attention block on abdominal MRI in \textbf{\textit{setting 1}}.}
\label{tab:block2}
\vspace{-0.5cm}
\end{table}

%% file: 5_Conclusion.tex
\section{Conclusion}
In this work, we propose a novel prototype-based method that
 introduces a novel Cycle-Resemblance Attention (CRA) module to fully leverage the pixel-wise relation between query and support medical images.
 In this way, pixel-wise spatial relationships between support and query images can be well preserved to address the loss of spatial information problem  in prototypical network. 
Overall, our network achieves a considerable improvement compared with the state-of-the-art approaches. On the abdominal-CT dataset, our method achieves over 10\% improvement on the mean dice score for all labels. Our ablation studies extensively illustrate our current implementation of different components to be optimized. Overall, our proposed method is intuitive and effective for the medical imaging semantic segmentation tasks with insufficient annotated data.

\noindent \textbf{Acknowledgments:} Y.Y. and D.C. received support by National Institutes of Health (NIH) 1RF1MH124611, 1RF1MH123402. D.C. also received support by National Science Foundation NeuroNex-1707316.

    